# Who Should Grade My Work? Student Perspectives on Transparent AI-Assisted Writing Assessment in Higher Education

**Rayed AlGhamdi**
Department of Information Technology
Faculty of Computing and Information Technology
King Abdulaziz University, Jeddah, Saudi Arabia
raalghamdi8@kau.edu.sa
ORCID ID: 0000-0002-6277-2124

## Abstract

The integration of Generative Artificial Intelligence (GenAI) tools into higher education assessment raises important questions about how students understand, interpret, and respond to AI-mediated evaluation. As instructors increasingly explore AI tools for providing feedback, prior research has examined whether GenAI-generated feedback improves writing performance and how students perceive its usefulness; comparatively little is known, however, about how students interpret such evaluation when they are explicitly informed that an AI system, rather than a human instructor, produced the feedback and the score. This study reports findings from a qualitative pedagogical inquiry conducted in an undergraduate technical communication course for computing students at a Saudi public university. Thirteen male undergraduate computing students completed an in-class handwritten writing task; the scanned submissions were evaluated by ChatGPT using a rubric-based prompt aligned with the task objectives. Students were then explicitly informed that ChatGPT had generated the score and feedback and were invited to reflect on the evaluation in writing. Inductive thematic analysis of these reflections identified four themes: perceived usefulness of feedback; awareness of AI's contextual and pedagogical limitations; conditional trust, distinguishing feedback utility from evaluative authority; and reflection on the institutional and pedagogical role of the human instructor. Participants accepted GenAI feedback as useful for surface-level revision but consistently positioned the human instructor as the appropriate authority over grading decisions. The study identifies this as a distinction between feedback utility and evaluative authority, two judgments that students treat as analytically separate rather than as opposite ends of a single approval scale. Within the bounded context and sample of this study, the findings suggest that transparency about GenAI involvement in assessment may shift student reflection from a narrow focus on feedback quality toward broader considerations of authority, trust, and assessment legitimacy. The paper proposes a three-principle framework (transparency, human mediation, reflective practice) for ethical GenAI integration in writing assessment.

**Keywords:** GenAI feedback; AI-assisted writing assessment; transparency in assessment; conditional trust; higher education

## 1. Introduction

Writing remains a central concern in higher education. Students who can articulate their ideas clearly are advantaged in coursework, professional communication, and employment outcomes (Hyland, 2019; AlGhamdi, 2019). For computing students in particular, written communication is important: graduates are expected to explain technical concepts to non-specialist audiences, document systems clearly, and justify design choices (ABET, 2025). The evaluation of student writing has, however, become more challenging in recent years. While concerns about the authenticity of student-submitted writing are not new, the rapid diffusion of generative AI tools since late 2022 has introduced a new dimension to the problem: students can now produce fluent drafts in seconds, complicating instructors' efforts to assess originality, effort, and how student work develops over time.

Empirical research indicates that such tools are now widely adopted by university students across disciplines, often as writing assistants rather than as reference aids alone (Balraj, 2025; Zhang et al., 2024). This raises a substantive challenge for instructors: assessing writing whose authorship and developmental process may be uncertain (Uddin et al., 2024; Delikoura et al., 2025). Prohibitive responses are of limited practical value given the accessibility of these tools to students; a more productive question concerns how instructors can engage with GenAI in ways that support, rather than displace, student learning.

One pragmatic response to this challenge is to involve GenAI in the provision of feedback. In large enrolment courses with weekly writing tasks, instructors often face practical limits on the depth and frequency of individualised feedback they can provide. Recent studies indicate that GenAI feedback may improve writing quality, increase student engagement, and reduce instructor workload when deployed thoughtfully (Kinder et al., 2025; Çağlar-Özhan et al., 2025). The same body of work, however, reveals mixed student responses, including scepticism about accuracy, concerns about generic feedback, and a persistent preference for human oversight (Aljasser, 2025; Le et al., 2025). Most existing studies in this area focus on writing performance outcomes (e.g., Kinder et al., 2025; Çağlar-Özhan et al., 2025) or on the perceived usefulness of AI feedback (e.g., Aljasser, 2025; Le et al., 2025), with comparatively less attention to how students interpret and make sense of GenAI evaluation once assessment authority and grading are involved.

One issue that has received less attention is transparency, specifically, whether students know that GenAI produced their evaluation. Research on algorithm aversion and trust suggests that perceptions of feedback quality and fairness are shaped not only by content but also by beliefs about its source (Dietvorst et al., 2015; Burton et al., 2020). Research on algorithm appreciation and aversion suggests that perceptions of source (algorithmic vs. human) influence acceptance of advice and decisions across domains (Logg et al., 2019; Castelo et al., 2019). Whether and how these source effects manifest in educational settings, particularly in writing assessment, remains less well established.

The present study was designed to examine how students respond when they know that AI has evaluated their work: how they appraise the feedback, whether and how they trust it, and how they assess its fairness. The study is positioned as part of a programme of pedagogical inquiry conducted within the same technical communication course. A prior study by the author was conducted under blinded conditions, in which students were unaware that ChatGPT had generated the feedback, so that their reflections were not shaped by knowledge of the source (AlGhamdi, 2024). At that time, student access to ChatGPT was relatively limited and few students were familiar with AI-assisted writing strategies. Direct causal comparison between the two studies is not possible, given differences in cohorts, time periods, and students' baseline AI familiarity; examining student responses under transparent conditions, however, offers a complementary perspective on how disclosure may shape engagement with AI-mediated evaluation. In the interval between the two studies, GenAI tools have become widely accessible and normalised in higher education, further complicating the evaluation of writing and the enforcement of originality (Zhang et al., 2024; Lee et al., 2025).

Accordingly, the present study adopts a transparent, reflection-oriented pedagogical design in which students are explicitly informed that ChatGPT evaluated their handwritten writing. Following the evaluation, students complete a short, in-class reflective task examining their agreement with the feedback and score, what the evaluation revealed about their writing, and their views on the appropriateness of using ChatGPT for writing assessment. By focusing on what students think rather than on the accuracy of GenAI feedback, this study addresses questions about trust, fairness, and learning that are central to responsible GenAI integration in higher education (Yusuf et al., 2024; Donadel and Zuin, 2026). In doing so, the study contributes a conceptual distinction that has not been clearly named in the existing literature: between students' acceptance of AI as a feedback mechanism (feedback utility) and their acceptance of AI as an evaluator with grading authority (evaluative authority). This distinction helps reconcile competing findings in the algorithm-trust literature and offers a more precise vocabulary for designing AI-mediated assessment practices.

### 1.1 Research Questions

This study addresses three research questions:

1. How do students evaluate ChatGPT-generated feedback once they know ChatGPT produced it?
2. How do students reason about ChatGPT's role as an evaluator of their work?
3. What broader pedagogical and institutional concerns emerge in students' reflections on AI-mediated assessment?

## 2. Literature Review

The emergence of GenAI tools in higher education has prompted a substantial body of empirical research examining how such tools function in writing pedagogy and how students respond to AI-generated feedback. This review situates the present study within several overlapping strands of that literature: research on the technical adequacy of AI feedback; research on student trust and algorithm aversion or appreciation; research on feedback literacy and the relational nature of assessment; research on reflective and metacognitive engagement with AI; research on ethical integration; and research on AI adoption in non-Western higher education contexts. Across these strands, a recurring observation is that the source of feedback shapes how students interpret and act on it, yet relatively few studies have examined what happens when students are explicitly told that an AI system, rather than a human instructor, has generated their evaluation.

### 2.1. Technical adequacy of GenAI feedback

A growing body of evidence indicates that GenAI feedback can match human feedback in important respects. Çağlar-Özhan, Tekeli, and Arkün-Kocadere (2025) compared AI-generated and instructor feedback directly and reported that students rated them similarly for clarity and usefulness when the AI was provided with a clear rubric, with no significant difference in perceived feedback quality or downstream performance. Kinder et al. (2025) reported comparable findings in a teacher education context, demonstrating that large-language-model feedback could be both adaptive and pedagogically aligned. These findings are reinforced by large-scale systematic reviews that conceptualise GenAI primarily as a tool for writing assistance, productivity enhancement, and pedagogical support rather than as a replacement for human instructors (Yusuf et al., 2024; Zhang et al., 2024).

This work establishes that technical adequacy is no longer the central question for AI feedback research. The question has shifted: not whether AI can produce useful feedback, but how students engage with it once they know it is AI. For the present study, this shift is important because participants were explicitly told that ChatGPT had evaluated their work, allowing the analysis to focus on the interpretive layer that sits above the technical content of the feedback itself.

### 2.2. Trust, algorithm aversion, and algorithm appreciation

A second strand of research examines how the source of advice or evaluation affects its acceptance. Dietvorst, Simmons, and Massey (2015) introduced the concept of *algorithm aversion*, demonstrating that people reject algorithmic decisions, even when algorithms outperform humans, particularly after observing them err. Burton, Stein, and Jensen (2020) extended this in a systematic review, identifying conditions under which algorithm aversion intensifies, including high-stakes decisions and domains perceived as requiring human judgement.

A counter-pattern, however, has also been documented. Logg, Minson, and Moore (2019) reported *algorithm appreciation*, showing that lay people often weight algorithmic advice more heavily than human advice in numeric estimation tasks. The same study and subsequent work suggest that algorithm appreciation is most pronounced in domains perceived as opaque or technical, and least pronounced in domains perceived as requiring distinctly human qualities, such as ethical judgement or character assessment. Castelo, Bos, and Lehmann (2019) similarly observed task-dependent algorithm aversion, finding that resistance to algorithms is stronger when tasks are perceived as subjective or as requiring uniquely human capacities.

Taken together, this body of work suggests that students may simultaneously appreciate AI for some functions (efficient, consistent, surface-level feedback) and resist it for others (high-stakes evaluation involving subjective judgement). The present study examines this distinction directly: rather than asking whether students accept or reject AI evaluation as a whole, the analysis attends to how students separate functions of AI within the same evaluation event.

### 2.3. Feedback literacy and the relational nature of assessment

A third strand of research, less prominent in the GenAI literature but central to the framing of the present study, concerns feedback literacy and the relational character of educational assessment. Carless and Boud (2018) defined student feedback literacy as the understandings, capacities, and dispositions needed to make sense of feedback and use it to enhance learning, organised around four inter-related features: appreciating feedback, making judgements, managing affect, and taking action. In a related earlier paper, Boud and Molloy (2013) contrasted two models of feedback: a transmission model derived from cybernetic and physiological metaphors, in which teachers act as the drivers of feedback, and a sustainable assessment model, in which learners take an active role in generating, soliciting, and using feedback for their own development.

Henderson, Ryan, and Phillips (2019), drawing on responses from 3,807 students and 281 educators in two Australian universities, identified three intersecting categories of feedback challenge: feedback practices themselves, contextual constraints such as workload and class size, and individual capacity for engagement with feedback. Their analysis frames feedback not as a discrete information transfer but as a socially constructed and contextually situated practice. Winstone and Carless (2019) developed this framing further, arguing that effective feedback design must shift attention from what teachers do to how students generate, make sense of, and use feedback.

These accounts are important for the present study for two reasons. First, they conceptualise feedback as a dialogical and interpretive process rather than as a one-way transmission, which informs how participants' reflections on AI evaluation can be read. Second, they highlight that students bring expectations about the assessment encounter, including expectations about who is evaluating them, what the evaluator can be expected to know about their work, and what dialogue is possible after the fact. When evaluative authority is delegated to an algorithmic system, these expectations are placed in question, but the existing feedback-literacy literature has not yet examined this specific case empirically.

## 2.4. Reflective engagement and metacognition with GenAI

A fourth strand of recent work has examined reflective and metacognitive dimensions of student engagement with GenAI tools. Lin et al. (2024) explored AI-assisted metacognitive strategies in academic reading and reported both cognitive and emotional tensions, with AI tools supporting some planning and evaluation functions while also imposing cognitive strain and exposing limitations in students' AI literacy. Xu, Su, and Liu (2024) examined student engagement with AI-augmented feedback in translation revision tasks, finding that students engaged selectively with AI suggestions and varied in their depth of cognitive engagement depending on task type. Chen et al. (2025) reported that students engaged differently with GenAI formative feedback in reading assessment, distinguishing between revising in response to feedback and merely receiving feedback as a recommendation. Yu, Wei, and Chen (2025) compared learners' engagement with AI chatbot feedback and peer feedback in an interpreter training programme, finding that learners were more likely to argue with AI feedback than with peer feedback, positioning AI as a negotiable rather than authoritative source. Across these studies, a consistent finding is that students engage with GenAI feedback selectively rather than as a single accept-or-reject decision, with task type, stakes, and source of feedback all shaping the depth and direction of engagement.

This strand of research is relevant to the present study because it indicates that GenAI feedback is not received passively even in formative contexts. Students interrogate, accept, dispute, and contextualise AI suggestions in ways that differ from how they engage with feedback from human instructors. The empirical question for the present study is whether and how this selective engagement intensifies when students are explicitly told that AI produced the evaluation, and when summative grading is at stake rather than formative revision.

## 2.5. Ethical concerns and the call for transparent integration

A fifth strand of work has addressed ethical concerns about AI in assessment, with growing attention to issues of authorship, academic integrity, and the legitimacy of automated evaluation. Systematic reviews caution that opaque or uncritical use of AI may contribute to superficial learning, over-reliance, and diminished student agency (Delikoura, Fung, & Hui, 2025; Lee, Huang, & Wu, 2025). Uddin et al. (2024) argued that prohibitive responses are neither realistic nor pedagogically productive given the accessibility of GenAI tools to students, and advocated instead for transparent, guided, and human-centred integration. Within such frameworks, transparency functions not merely as disclosure but as a pedagogical strategy that surfaces learners' assumptions, concerns, and values regarding GenAI.

The empirical question of what transparency about AI involvement actually does for student engagement, however, remains under-examined. Aljasser (2025) compared instructor and ChatGPT-generated feedback in a Saudi EFL context and reported student appreciation for AI feedback alongside continued preference for human oversight. Le et al. (2025) examined learner preferences for feedback from GenAI versus human tutors and reported nuanced patterns rather than uniform preference. Donadel and Zuin (2026) explored student dispositions and experiences with GenAI in higher education through a mixed-methods design and found heterogeneous patterns of trust, scepticism, and conditional acceptance. The present study contributes to this strand by examining a transparent, post-assessment reflective design in which students know an AI evaluated their work, and asking what they reason about under those conditions.

### 2.6. GenAI research in non-Western higher education contexts

Research on GenAI in higher education has been concentrated in Western institutional contexts, with comparatively less empirical attention to Middle Eastern, African, and parts of Asian higher education (Yusuf et al., 2024; Zhang et al., 2024). This concentration matters because cultural and institutional factors shape how students reason about evaluative authority. Saudi higher education sits within a broader cultural context characterised in cross-cultural research as relatively high in power distance and collectivism (Hofstede, 2011), in which the instructor is often positioned not only as an evaluator but also as a respected figure embedded in institutional and pedagogical structures. While Hofstede's framework has been subject to critique and refinement in more recent scholarship, the broader observation that student-teacher relationships in Saudi higher education carry strong relational and institutional weight is consistent with the contextual literature (Sobaih, Elshaer, & Hasanein, 2024; Alsofyani & Barzanji, 2025).

Empirical work on ChatGPT in Saudi higher education has grown rapidly. Sobaih, Elshaer, and Hasanein (2024) examined ChatGPT acceptance among 520 Saudi university students through the UTAUT2 framework and reported broad acceptance alongside concerns about reliability and academic integrity. Alsofyani and Barzanji (2025) reported that ChatGPT-generated feedback was as effective as teacher-generated feedback for Saudi EFL writers, with generally positive student perceptions, while also surfacing concerns about appropriate use. These studies establish a baseline of student receptivity to AI in Saudi contexts but examine acceptance rather than reflective engagement with AI evaluation as such.

The present study contributes to this gap by examining how Saudi computing students reason about AI evaluation when they know an AI evaluated their writing, attending to the categories of reasoning that emerge from their reflections rather than to predefined acceptance constructs.

### 2.7. Summary and gap

Taken together, the literature reviewed above points to four principal observations. First, AI feedback has been shown to match instructor feedback in clarity and instructional utility for many writing tasks. Second, students engage selectively with such feedback, expressing nuanced and sometimes contradictory reservations regarding accuracy, fairness, and evaluative authority, particularly when grades are at stake. Third, feedback is not received as a discrete information transfer but is interpreted within a relational understanding of who is evaluating, what they can know about the learner, and what dialogue is possible after the evaluation. Fourth, transparency about AI involvement in assessment appears to shape student perceptions, but the specific mechanisms and consequences of disclosure, particularly in non-Western higher education contexts, remain under-examined.

The present study addresses this gap by examining, in a transparent post-assessment reflective design, how Saudi computing students reason about AI-mediated writing evaluation when they have been explicitly informed that an AI system generated their feedback and score. The analysis attends not only to whether students accept or reject AI feedback, but to the distinctions students themselves draw between different functions of AI within the assessment process.

## 3. Methodology

### 3.1. Research Design

This study emerged from a practitioner-researcher question: how do students respond when the instructor is transparent about using AI to evaluate their work? The study adopted a qualitative, course-embedded pedagogical inquiry approach, using inductive thematic analysis of student reflective writing. The setting was an undergraduate Technical Communication course taught by the author, in which the aim was to examine how students respond when explicitly informed that ChatGPT, rather than the instructor, has evaluated their writing. Unlike the author's earlier blinded study (AlGhamdi, 2024), in which students were unaware that ChatGPT had generated the feedback, this study foregrounds transparency by positioning GenAI evaluation as an explicit instructional intervention rather than a concealed assessment mechanism.

The study is framed as a course-embedded pedagogical inquiry, responding to the contemporary challenge that widespread student access to GenAI has altered writing practices, authorship, and assessment authenticity in higher education (Zhang et al., 2024). Rather than attempting to prohibit GenAI use, the current study examines how explicit AI-mediated evaluation may reshape students' metacognitive awareness, trust, skepticism, ethical reasoning, and engagement with feedback.

### 3.2. Context and Participants

The study was conducted at a Saudi public university, within a Technical Communication course offered to undergraduate students at a computing faculty. The course emphasises practical written communication skills, including clarity of expression, logical organisation, and audience awareness.

The study sample consisted of 19 male students enrolled in a single section of the course during the study semester. All participants were in their second year at the university level, having completed a foundation year before joining the computing faculty. As part of routine coursework, students completed a weekly writing task followed by structured feedback and scoring. The reflective component analysed in this study was embedded within regular assessment activities rather than administered as a separate research instrument. Of the 19 enrolled students, 13 completed the reflective task and were included in the final analysis. The analytic sample, therefore, represents those students who submitted complete reflective responses following the AI-generated evaluation.

To provide a clear overview of the research process, Figure 1 illustrates the six-stage methodology employed in this study, from the initial writing task through to thematic analysis.

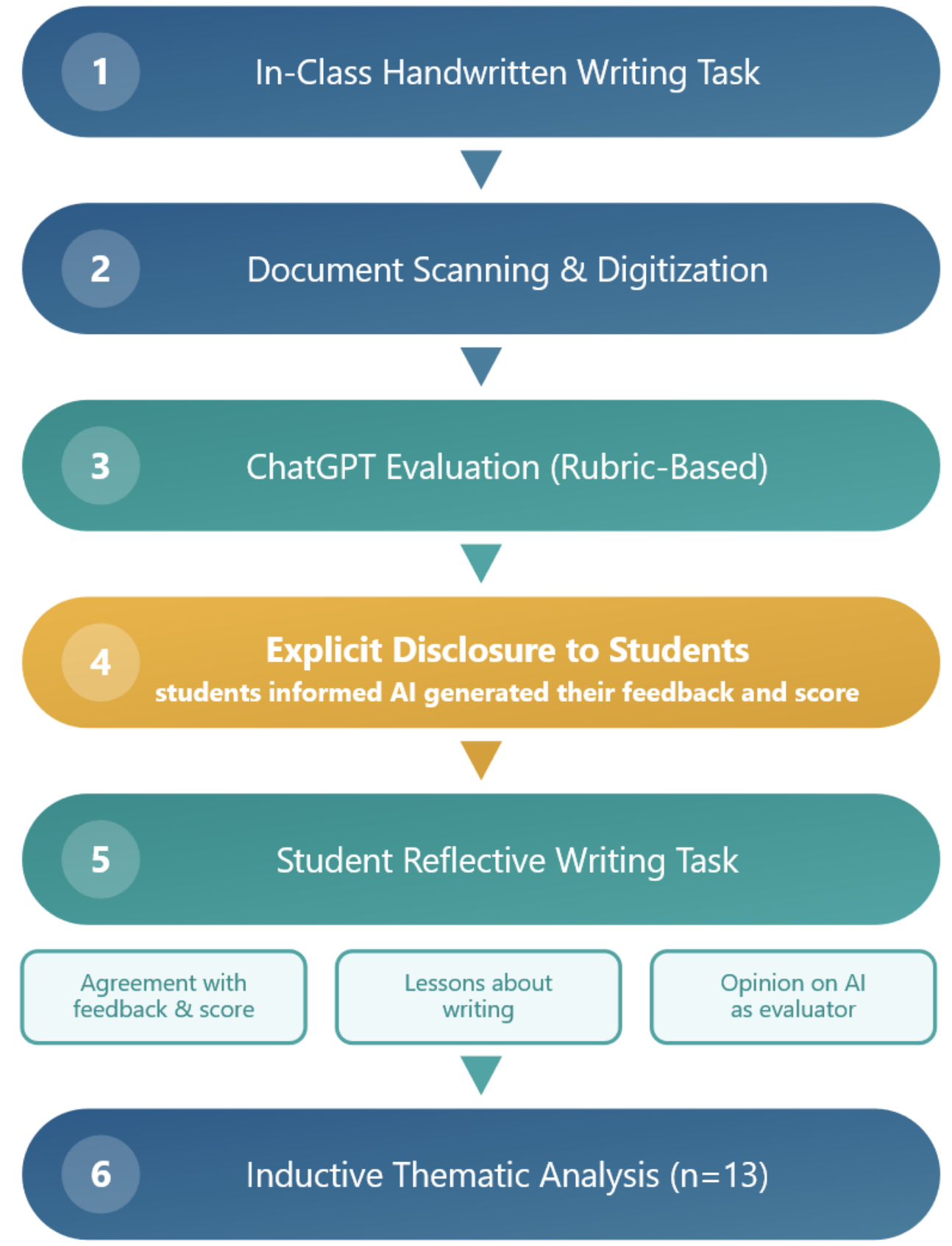

Figure 1. Study procedure. In-class handwritten writing tasks (1) were scanned (2) and evaluated by ChatGPT against the course rubric (3). Students were then explicitly informed that an AI had generated their feedback and score (4), and were invited to complete a reflective writing task (5) addressing three prompts: agreement with the feedback and score, lessons drawn about their writing, and opinions about AI as an evaluator. The 13 returned reflections were analysed using inductive thematic analysis (6).

As shown in Figure 1 above, the design centres on transparency as the key intervention (Stage 4), distinguishing this study from conventional AI-assisted feedback approaches. Students were explicitly informed that ChatGPT evaluated their handwritten work before completing the reflective task.

The sample size aligns with established benchmarks for reflexive thematic analysis with homogeneous bounded samples (Guest et al., 2006; Malterud et al., 2016; Braun & Clarke 2021), where 9 to 17 participants is commonly considered sufficient for analytic saturation.

ChatGPT, as a GenAI tool, was selected for this pedagogical activity for three reasons. First, it represents the GenAI tool most familiar to students in the local higher education context at the time of the study, and therefore aligns with the assessment ecology that students themselves encounter. The study aimed to examine students' perceptions of AI evaluation within a realistic context of contemporary academic practice, rather than to compare specific AI platforms. Second, ChatGPT's instruction-following capabilities permitted the formulation of structured evaluation prompts aligned with the course rubric (clarity, organisation, sentence correctness, conciseness), enabling consistent, criterion-referenced feedback across submissions. Third, at the time of the study, no institutionally governed AI evaluation platform was available at the host institution for this purpose. The author acknowledges that other tools, including institutionally integrated AI assistants, may offer enhanced data-protection guarantees and that institutional policies on AI tool selection are evolving rapidly. Future iterations of this research could productively compare student perceptions across different AI platforms to determine whether the patterns observed here reflect responses to AI evaluation in general or to ChatGPT specifically; this is acknowledged as a limitation in Section 6.

In-class handwritten writing was chosen deliberately as a methodological feature rather than as a technological constraint. Given widespread student access to GenAI tools at the time of the study, handwritten in-class production assured that the writing being evaluated was generated by the student, not by an AI system. This design choice was essential to the integrity of the study: examining student perceptions of AI evaluation requires that the writing under evaluation be authentically the students' own. Handwritten production also created the conditions for examining how students respond when an AI system evaluates work that is unambiguously theirs, foregrounding the questions of trust, authority, and assessment legitimacy that this study set out to investigate.

In the first stage of the study, students completed an in-class writing task by hand. Figure 2 presents a verbatim transcription of an anonymised student handwriting submission. These handwritten submissions were scanned and provided to ChatGPT, which evaluated the writing, generated feedback, and assigned a score that counted toward grades based on predefined criteria aligned with course objectives (clarity, organization, sentence correctness, and conciseness).

Task prompt (printed):

Reflect on your MakeMeBuy experience. Describe your first impression when you read the task instructions and when you watched the samples from previous students. Explain how you planned and created your own video, how you felt during the process, and how you felt after completing it. Mention what you liked most about the task, the challenges you faced, and how you overcame them. Describe how you shared your video with others and the kind of feedback you received. Finally, explain what lessons you learned from this whole experience.

Student response (handwritten):

When i read the instructions, I wasn't sure about how the video should be, I watched the student samples and i noted that Dr. [instructor] gave the students complitly freedom of showing their content with the way they want.

~~So~~ I started searching about uniqe products, I found many but at the end i decided to choose Starlink.

I wrote a script, set the camera, and start recording. Everytime i record a scense. I reapeted about twenty times, this took me so much time, by the time pass i set my confidence on, that helps me with reducing my records per scence.

Figure 2. Verbatim transcription of an anonymised student handwriting submission subsequently evaluated by ChatGPT. Student identifiers redacted and instructor name replaced with [the instructor]. Original spelling, grammar, and self-corrections preserved for analytic transparency.

Figure 3 illustrates an example of ChatGPT-generated feedback and score provided to students for their handwritten writing. Students were explicitly informed that ChatGPT evaluated their handwritten work.

Score: ⭐ 1.5 / 2 points

**Evaluation:**

- **Clarity (0.5/0.5):** The reflection clearly expresses excitement, effort, and learning. The message is understandable and personal.
- **Organization & Structure (0.25/0.5):** The ideas are mostly chronological, but transitions between sentences need smoother connection. Some repetition and skipped details (e.g., how challenges were solved) make it feel slightly unorganized.
- **Sentence Correctness (0.5/0.75):** Grammar and spelling need some editing ("so much value in me" → *gave me so much value*, "it let you so what ever you want" → *it let you do whatever you want*). However, meaning is still clear.
- **Conciseness (0.25/0.25):** The writing stays focused and brief.

**Feedback for Improvement:**

Good effort, ! You shared your enthusiasm and what you learned clearly. To improve, work on **sentence connections** and **grammar accuracy**. Try reading your writing aloud after finishing — it helps you notice where sentences sound disconnected or unclear. Add a short reflection on *how* you solved your challenges to make your story stronger.

Figure 3. Example of ChatGPT-generated feedback & score provided to students for their submissions

After receiving the ChatGPT-generated evaluation and score, students were asked to complete a reflective writing task responding to three prompts:

1. Whether they agreed or disagreed with the AI feedback and score, and why
2. What they learned or noticed about their writing from the evaluation
3. Their opinion about using ChatGPT to evaluate writing and provide feedback

### 3.3. Data Collection

The primary data source consisted of students' handwritten reflective responses, produced during a structured in-class reflective task that followed receipt of the ChatGPT-generated evaluation. In this task, students were provided with (a) their original handwritten submission and (b) the ChatGPT-generated feedback and score. They were asked to reflect on the evaluation using pen and paper.

These handwritten reflections were spontaneous, low-stakes responses that were intended to reduce performative bias and capture authentic student reactions. The handwritten reflections constitute the sole dataset analyzed in this study. Figure 4 below demonstrates a sample of handwritten student reflections in response to ChatGPT-generated feedback.

Task prompt (printed):

You received feedback and a score from ChatGPT on your most recent writing task. Reflect on that evaluation and share your opinion about it.

- Whether you agree or disagree with the feedback and score, and why.
- What did you learn or notice about your writing from the evaluation?
- Your opinion about using ChatGPT to evaluate your writing and give writing feedback.

Student response (verbatim):

I agree with chatGPT, but I think The gread must be writing by doctor not chatGPT. The not[e]s was clear and the gread was nice. But if chatGPT cheaking the exams why we are going to univecity? Why not learn from chatGPT and test ourself in chatGPT? The doctors shud cheaking the student exam and gread it. Ai may has mestak and this is not good for students. Doctors can use Ai for read it feedback for exams only, but if they use it for gread students problems will hapend.

Figure 4. Verbatim transcription of an anonymised student reflection in response to ChatGPT-generated feedback, illustrating the type of reflective data analysed in this study. Original spelling, grammar, and self-corrections are preserved; student identifiers have been redacted.

### 3.4. Data Analysis

To analyze students' reflections, an inductive thematic analysis was conducted following Braun and Clarke's (2006; 2021) guidelines. The analysis was data-driven, allowing themes to emerge from students' own words rather than from predetermined theoretical categories. The process involved: (a) repeated familiarisation with the dataset; (b) open semantic coding of meaningful units related to perceptions of ChatGPT feedback, trust and skepticism, learning awareness, emotional responses, ethical concerns, and comparisons with human evaluation; and (c) iterative grouping and refinement of codes into coherent themes through constant comparison. Final themes were clearly defined and analytically described with attention to their pedagogical relevance to AI-mediated writing assessment. Representative student quotations were retained to ground analytic claims, and reflexivity was maintained by centering analysis on students' self-articulated experiences rather than on evaluative judgments of writing quality.

The thematic analysis was conducted in two complementary stages. In the first stage, the author conducted the primary coding using NVivo, working inductively from the transcribed reflections to identify open codes, group these into provisional themes, and refine the theme structure through iterative engagement with the data. The four themes reported in Section 4 emerged from this primary analysis. In the second stage, the scanned handwritten reflections were independently re-analysed using Claude Code (Anthropic, model Opus 4.7) as a confirmatory analytic check, conducted with no prior access to the author's thematic structure. This sequencing, with human-led coding first and AI-assisted verification second, was adopted deliberately to ensure that the author's analytic judgement was not anchored by AI-generated framings (AlGhamdi, 2026).

The confirmatory analysis converged with the primary analysis on Themes 1 and 2 and prompted refinement of the framings used for Themes 3 and 4 to more closely reflect participants' own articulations: Theme 3 was sharpened to highlight the distinction between feedback utility and evaluative authority, and Theme 4 was reframed from a focus on transparency and fairness to a focus on participants' reasoning about the institutional and pedagogical role of the human instructor. These refinements were adopted on the basis of re-examining the data, not on the basis of accepting the second analyst's interpretations uncritically. The use of AI-assisted analysis is acknowledged as having limitations, including the possibility of shared interpretive biases between large language models; future replication studies would benefit from independent second coding by a human analyst.

### 3.5. Ethical Considerations

Several specific procedures warrant explicit discussion here. First, regarding data handling, handwritten student submissions were scanned and the resulting images were processed through ChatGPT for evaluation; all submissions were anonymised before processing, with student names and any identifying information removed or redacted, and no personally identifying metadata was retained in the dataset analysed for this study. Second, regarding student autonomy, the AI-mediated evaluation was implemented as a course-level pedagogical activity, not as a research-specific procedure imposed on individual students; students who preferred not to have their work used for the research component were free to decline, and the use of their reflections for research purposes was distinct from their participation in the routine course activity. No student declined to participate in research, and no identifiers were linked to the analytic dataset. Third, regarding the dual role of instructor-researcher, this position introduces well-recognised risks of response bias and power imbalance; the study mitigated these risks through (a) framing the reflective prompt as open-ended inquiry that explicitly welcomed critical and divergent views, (b) embedding the reflection within routine coursework rather than presenting it as a research instrument, (c) anonymising responses prior to analysis, and (d) centring analytic attention on students' self-articulated experiences rather than evaluative judgements of writing quality. The author acknowledges that residual bias cannot be entirely eliminated under such conditions, and notes this as a limitation in Section 6.

While the instructor-researcher position carries the risks noted above, it also permitted contextually grounded insight into how students engaged with feedback in this course. During analysis, the author focused on understanding the meaning of students' statements rather than evaluating their correctness; doubts and disagreements were treated as analytically meaningful data rather than as inconsistencies to be resolved.

## 4. Findings

This section reports the findings from inductive thematic analysis of students' written reflections on ChatGPT-generated evaluations of their weekly writing tasks. The analysis examined how students perceived the evaluation when they were explicitly informed that ChatGPT had generated the feedback and the score. Four themes emerged consistently across the dataset: (1) perceived usefulness of feedback; (2) awareness of AI's contextual and pedagogical limitations; (3) conditional trust, distinguishing feedback utility from evaluative authority; and (4) reflection on the institutional and pedagogical role of the human instructor. Table 1 summarises the distribution of these themes across the thirteen participants. The themes are presented separately for analytic clarity, although in students' reflections they were often intertwined within a single response.

### 4.1. Perceived usefulness of feedback

All thirteen participants described the ChatGPT-generated feedback as useful, particularly valuing its identification of specific issues, for example, a grammar error or a disorganised paragraph, rather than vague evaluative comments such as "needs improvement." Several students described the feedback as "clear," "direct," "well-organized," and "objective." One participant wrote: "The comments were clear and helped me understand my mistakes, especially in grammar and sentence structure." Another stated: "[ChatGPT] showed my mistakes directly and gave me specific points to improve. It also made me more aware of how to write in a clearer and more organized way." A third participant described the feedback as "very helpful" because "it gave me clear specific feedback points, especially the suggestion to work on grammar and spelling." This universal endorsement of feedback usefulness is itself a noteworthy finding, given the contested nature of student responses to AI in other studies (see Section 5.1).

Several participants explicitly identified what they had learned from the feedback. One participant noted that the evaluation "helped me build confidence in my writing while also providing advice to improve the writing even more." Another stated that the feedback "explained clearly what mistakes I avoided that the average student would make, while pointing out the unique traits seen from my writing," adding that "the scores matched the explanation perfectly." These responses indicate that participants recognised the technical value of ChatGPT feedback, particularly for surface-level writing features. The consistency and systematic nature of the feedback were highlighted as strengths, especially when compared with the brief or delayed feedback often encountered in large classes.

### 4.2. Awareness of AI limitations

While participants valued the feedback, the majority also articulated specific limitations of the AI evaluation. These limitations fell into two categories: technical and contextual. At the technical level, several participants raised concerns that ChatGPT could misinterpret handwriting, writing intent, or contextual meaning, especially because the writing was handwritten and later scanned. One participant observed: "I did notice that it spelt my last name wrong and it thinks I made mistakes in my spelling when I didn't. I think this is because of my handwriting so it must have misread it." This participant added: "I would also take its score less seriously because, as I mentioned, it can misread words and possibly make false assumptions." Another stated more bluntly: "AI can make mistakes that will affect the student's grades."

At the contextual level, participants identified limitations that went beyond technical error. One participant pointed to a misalignment between the AI's evaluation and the institutional rating system: "The AI rating had one problem, which is the AI doesn't know [the instructor's] rating system." Another raised a concern about AI's consistent positivity: "My opinion about using AI to give feedback is [that] AI always tries to be positive, but that doesn't mean it's unable to use. I think we can use it as initial feedback." These limitations concern AI's capacity to interpret context, intent, effort, and institutional norms rather than its capacity to produce fluent text.

These responses suggest that at least some participants in this study did not treat ChatGPT evaluation as neutral or infallible. Instead, they actively questioned its accuracy and reliability, especially regarding contextual understanding and the interpretation of meaning.

### 4.3. Conditional Trust: Distinguishing Feedback Utility from Evaluative Authority

The most analytically distinctive pattern in the dataset was that participants could simultaneously affirm the usefulness of ChatGPT's feedback and reject ChatGPT as an appropriate final evaluator. Participants drew a consistent distinction between "this feedback is useful" and "this system should decide my grade." These were treated as analytically separate judgements rather than two ends of a single approval scale. The pattern was articulated in many forms across the dataset. One participant captured the distinction concisely: "In my opinion on ChatGPT feedback, I agree with it, but with one condition, that the doctor checks the feedback to edit any mistakes." Another formulated the same position by reference to the instructor's superior knowledge of students: "The teacher knows much better about students and knows better [about] good ways to grade students and support students' learning.".

Some participants explicitly endorsed AI for one role while rejecting it for another. One wrote: "Doctors can use AI for [reading] feedback for exams only, but if they use it for [grading] students, problems will happen." Another distinguished between "initial feedback" and final grading, accepting the former and questioning the latter. A particularly articulate participant grounded the rejection of AI-as-grader in dialogical reasoning: "ChatGPT is not like people. You talk to them and explain excuses to consider or give another chance if you fail. People talk and discuss problems and understand each other's problems. … I like normal teacher [to] grade my work." Notably, the same conditional-trust pattern was observed even among participants most enthusiastic about ChatGPT: one who reframed ChatGPT as "your own teacher" nevertheless framed the endorsement as relating to learning support rather than grading authority, and another, initially "skeptical," recommended that "more teachers should at least try using AI, especially for feedback", explicitly limiting the endorsement to feedback rather than evaluation.

### 4.4. Reflection on the Institutional and Pedagogical Role of the Human Instructor

Several participants moved beyond an appraisal of the feedback itself to articulate reasons why the human instructor matters in assessment. These reflections went beyond technical limitations of AI to address the pedagogical and institutional functions that, in participants' view, only a human teacher can fulfil. Four interrelated arguments appeared across the dataset. First, participants invoked the teacher's knowledge of individual students. One participant wrote: "Teacher knows much better about students and knows good ways to grade students and support students' learning." Another participant, writing partly in Arabic at the end of the reflection, elaborated that there are aspects of teaching that ChatGPT cannot replicate, such as engaging with students individually and explaining concepts in personally appropriate ways.

Second, participants emphasised the dialogical nature of human assessment: "ChatGPT is not like people. You talk to them and explain excuses to consider or give another chance if you fail." This argument framed grading not as a one-way judgement but as a relational practice that requires interpretation and the possibility of dialogue. Third, participants raised institutional concerns about the role of the university itself when AI takes over evaluation. One participant articulated this in pointed terms: "If ChatGPT [is] checking the exams, why are we going to university? Why not learn from ChatGPT and test ourselves in ChatGPT?" This question is not primarily about feedback quality; it concerns the legitimacy and purpose of formal higher education when assessment is delegated to AI. Fourth, participants raised fairness-related concerns specifically connected to the consequences of AI error: "AI can make mistakes that will affect the student's grades." Such concerns positioned the instructor as a necessary safeguard against the propagation of AI errors into consequential evaluative decisions. Together, these arguments construct a pedagogical case for instructor authority that is distinct from, and stronger than, simple opposition to technology. Participants were not rejecting AI; they were articulating positive reasons why human evaluative authority should be preserved in higher education assessment.

### 4.5. Summary of Findings

Overall, the findings suggest that participants in this study:

1. Recognized the technical usefulness of ChatGPT-generated feedback
2. Demonstrated critical awareness of AI limitations
3. Articulated a distinction between feedback utility and evaluative authority, accepting AI in the former role while preferring human authority in the latter
4. Reasoned about the institutional and pedagogical role of the human instructor as a safeguard in AI-mediated assessment

These patterns emerged from a small, purposive sample and should not be generalized without further research.

Table 1. Distribution of themes across participants (N = 13)

| Participant | T1: Perceived Usefulness | T2: Awareness of Limitations | T3: Conditional Trust | T4: Instructor Role |
|---|---|---|---|---|
| P1 | ● | ◐ | ● | ● |
| P2 | ● | ● | ◐ | |
| P3 | ● | ● | ● | ● |
| P4 | ● | | | |
| P5 | ● | ● | ● | |
| P6 | ● | | | |
| P7 | ● | | | ◐ |
| P8 | ● | ◐ | ◐ | ● |
| P9 | ● | | ◐ | |
| P10 | ● | ● | ● | |
| P11 | ● | ● | ● | ● |
| P12 | ● | ● | ● | ● |
| P13 | ● | ● | ● | ◐ |
| **Total** | 13 | 7 (+2) | 7 (+3) | 5 (+2) |

Note: ● = theme explicitly articulated; ◐ = theme implicit or partially articulated; blank = theme not articulated. The Total row indicates the count of participants with explicit articulation, with the number of additional partial articulations in parentheses where applicable.

See also figure 5.

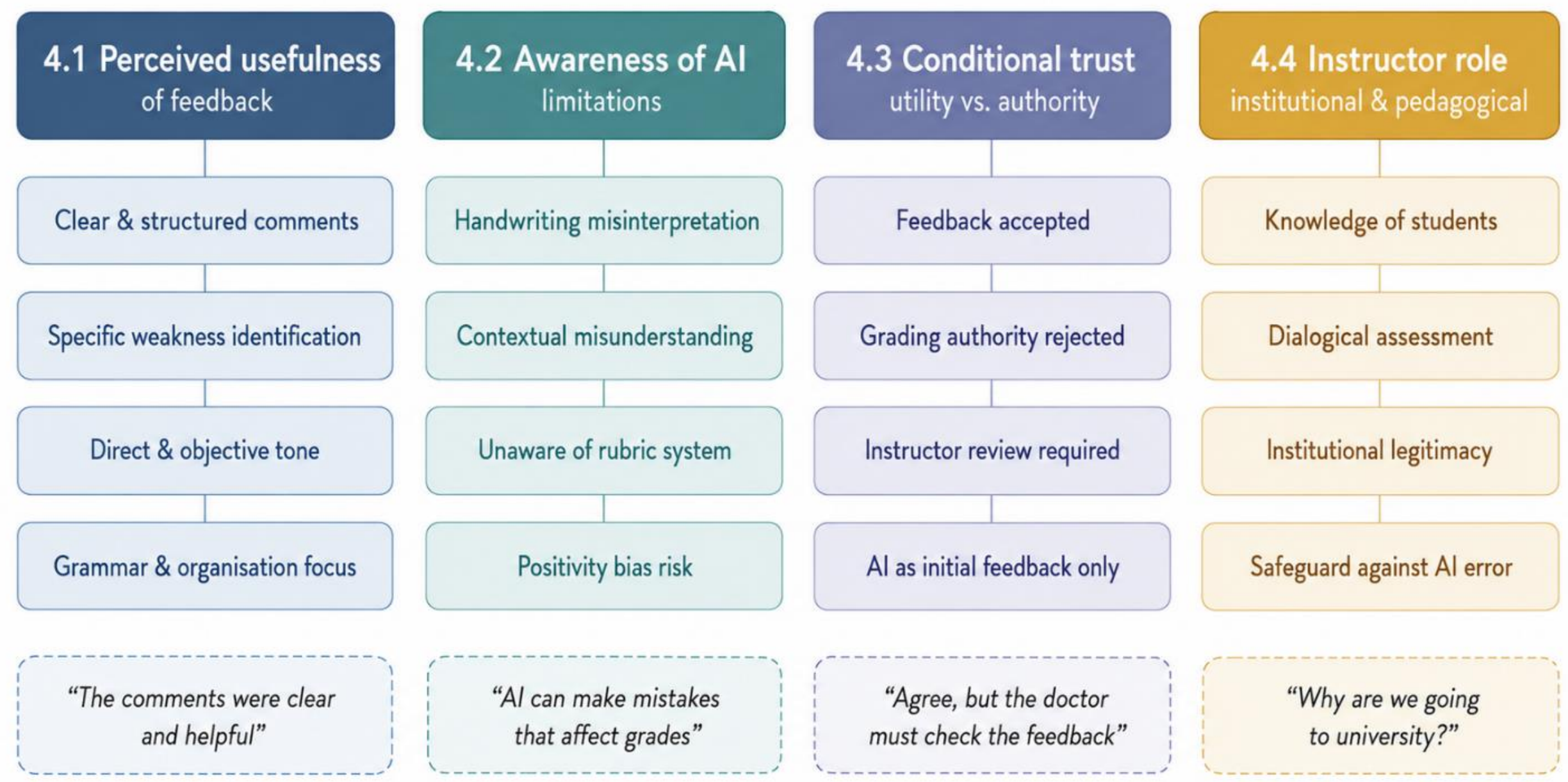

Figure 5. Thematic coding map of the four themes identified through inductive thematic analysis of student reflections (N=13), showing representative sub-themes and illustrative participant quotations for each theme

These findings provide the empirical foundation for the discussion that follows.

## 5. Discussion

A central observation across the dataset was that when students were aware ChatGPT had evaluated their work, their reflections extended beyond appraisal of the feedback itself to broader questions about the assessment process. Participants raised questions about fairness, the appropriateness of machine-generated grades, and the instructor's role in AI-mediated assessment. These concerns extended beyond whether the feedback content was accurate.

The subsections that follow discuss the four interrelated patterns in detail, situating the findings within existing research on AI-mediated feedback, transparency, and writing pedagogy. A conceptual model synthesising these patterns is presented at the end of the section (Figure 6).

### 5.1. Distinguishing Feedback Utility from Evaluative Authority

One prominent pattern observed in this sample was the distinction between students' agreement with AI-generated feedback and their acceptance of AI as an authoritative evaluator. While many students acknowledged the clarity, structure, and technical usefulness of the feedback provided by ChatGPT, this recognition did not translate into unconditional trust in AI-based grading.

Students' reflections indicate that ChatGPT feedback was primarily valued for its ability to identify surface-level issues in writing, such as grammar, sentence construction, organization, and conciseness. These perceptions align with previous research showing that GenAI can provide feedback comparable to human evaluators for lower-order writing concerns (Çağlar-Özhan et al., 2025; Kinder et al., 2025). In this study, transparency in AI evaluation appeared to activate a more critical stance, prompting students to separate feedback utility from assessment legitimacy.

Rather than accepting AI as a neutral or superior assessor, participants consistently emphasised that grading carries institutional and ethical weight that should remain the responsibility of a human instructor. This position aligns with the well-documented phenomenon of algorithm aversion: the tendency of individuals to reject algorithmic decisions even when algorithms outperform humans, particularly in domains where human judgement is perceived as essential, such as education (Dietvorst et al., 2015). Algorithm aversion has been shown to intensify as the stakes of the decision increase (Burton et al., 2020), consistent with the strong preference for instructor oversight expressed by participants in the present study when grading was at issue. Participants' scepticism did not appear to stem from a rejection of technology itself, but rather from awareness of AI's contextual limitations and its lack of pedagogical accountability.

The pattern observed in this sample also helps to reconcile competing findings in the broader literature. While Dietvorst et al. (2015) and Burton et al. (2020) document algorithm aversion in stakes-bearing decisions, Logg, Minson, and Moore (2019) reported algorithm appreciation in numeric estimation tasks where lay people weighted algorithmic advice more heavily than human advice. Castelo, Bos, and Lehmann (2019) demonstrated that this aversion is task-dependent, intensifying when tasks are perceived as requiring uniquely human capacities such as subjective judgement. Participants in the present study expressed neither generalised aversion nor generalised appreciation; instead, they distinguished between AI functions within the same evaluation event, accepting AI for surface-level feedback while preserving human authority for evaluation. This task-specific pattern of trust adds nuance to both algorithm-aversion and algorithm-appreciation accounts, suggesting that students are not making a single accept-or-reject decision about AI but are instead making separate judgements about distinct functions of the same system.

An earlier study in the same course, conducted under blinded conditions with a different cohort, found that students tended to focus on feedback content without questioning its source or authority (AlGhamdi, 2024). In contrast, the current cohort, who knew ChatGPT was involved, explicitly articulated boundaries around AI's role. While these studies cannot be directly compared (different students, different time periods, different levels of AI familiarity in the broader culture), the contrast is suggestive: transparency may have helped transform feedback from a taken-for-granted instructional act into an object of reflection. However, this interpretation remains tentative and requires verification through within-subject or controlled experimental designs.

## 5.2. Conditional Trust and Instructor Oversight

Among participants in this study, trust in ChatGPT-generated feedback appeared to be conditional. While students acknowledged the efficiency and clarity of ChatGPT's feedback, this trust was consistently framed as contingent upon instructor presence, validation, and oversight. Rather than viewing AI as an independent evaluator, students positioned it as a supplementary tool whose outputs require human interpretation and contextualization.

Comparable patterns have been reported in studies showing that students engage selectively with AI suggestions, accepting some and disputing or ignoring others (Chen et al., 2025; Yu et al., 2025). In the present study, awareness that an AI system was the evaluator appeared to intensify this selective engagement: participants did not accept feedback uncritically simply because it had been algorithmically generated. Transparency may therefore have encouraged participants to actively interrogate the feedback process, leading them to articulate expectations regarding accountability, fairness, and pedagogical responsibility.

This pattern of selective, conditional trust extends prior work on the relational character of feedback. Boud and Molloy (2013) argued that feedback should be conceptualised not as a discrete information transfer but as a sustainable practice in which learners actively engage with feedback within a relational context. Carless and Boud (2018) similarly framed student feedback literacy as a capacity that includes making judgements about feedback rather than receiving it passively. Henderson, Ryan, and Phillips (2019) further emphasised that feedback is socially constructed and contextually situated rather than a discrete transfer of information. Participants in the present study exhibited precisely this capacity for active judgement, but extended it in a direction that has received less empirical attention: they distinguished not only what feedback to accept but also which source has the authority to evaluate. This extension suggests that in AI-mediated contexts, feedback literacy includes an additional component, the capacity to reason about evaluative authority itself, separate from the capacity to reason about feedback quality.

Participants' emphasis on instructor oversight also reflects broader ethical concerns documented in the literature, especially regarding the risks of delegating high-stakes academic decisions to automated systems (Uddin et al., 2024; Delikoura et al., 2025). Students expressed concern that AI systems may misinterpret context, overlook intent, or inadequately account for individual effort. These were limitations that they believed necessitated human judgment. These concerns did not appear to reflect resistance to innovation but rather seemed to represent calls for balanced integration in which GenAI augments, rather than replaces, instructor expertise.

This pattern, observed in a Saudi computing course, is consistent with broader Saudi-context findings of student acceptance of AI feedback alongside continued preference for human oversight (Aljasser, 2025; Alsofyani and Barzanji, 2025; Sobaih, Elshaer, and Hasanein, 2024), and extends those findings by identifying the specific distinction between feedback utility and evaluative authority that participants drew.

## 5.3. Pedagogical Effects of Transparency

Explicitly informing students that ChatGPT had evaluated their writing appeared to function as more than a procedural disclosure; it operated as an instructional intervention in its own right. Once students were aware that an AI system had generated the evaluation, their reflections moved beyond engagement with the feedback content to a broader examination of who holds authority to judge their work, what trust in evaluation requires, and what role human instructors should play in AI-mediated assessment. By explicitly disclosing that ChatGPT generated the evaluation, the assessment process was reframed from a routine instructional practice into an object of reflection and inquiry.

This pattern is consistent with research suggesting that transparency about GenAI involvement can activate learners' metacognitive and ethical awareness by making otherwise invisible systems open to scrutiny (Zhang et al., 2024; Delikoura et al., 2025). In the present study, transparency appeared to prompt students to question assumptions about objectivity, accuracy, and fairness, particularly in relation to AI's capacity to interpret context, intent, and effort.

Notably, transparency did not lead to disengagement or resistance. Instead, students adopted a reflective stance in which agreement with feedback was accompanied by evaluation of its limitations. Students articulated concerns about over-reliance on AI, the risk of misinterpretation, and the necessity of human judgement, demonstrating a level of ethical reasoning that extended beyond immediate performance outcomes. This finding resonates with literature emphasising the role of reflective practices in fostering feedback literacy (Carless and Boud, 2018) and responsible GenAI use in education (Uddin et al., 2024; Delikoura et al., 2025).

For practitioners, this finding has an immediate implication: transparency about AI involvement may be more than an ethical obligation. It may be a pedagogical strategy in its own right, one that prompts students to engage critically with feedback rather than accepting it passively. Instructors considering AI-assisted feedback might therefore frame disclosure not as a disclaimer but as an instructional prompt.

### 5.4. Post-Assessment Reflection and System-Level Awareness

Most research on GenAI and writing focuses on reflection during drafting or revision. This study instead asked students to reflect after receiving their scores, a deliberate choice. This temporal positioning proved consequential. By inviting students to reflect after receiving a GenAI score and feedback, the task shifted students' attention beyond textual improvement toward a broader examination of the assessment system itself.

Students' reflections moved from questions of 'How can I improve this sentence?' to more systemic considerations such as 'Who should evaluate my work?' and 'What role should AI play in grading?', suggesting that the temporal positioning of reflection, after rather than during evaluation, shifted the focus of student reasoning toward broader questions about the assessment system.

This post-assessment reflection did not appear to diminish students' engagement with writing quality. Instead, it reframed writing as part of a larger evaluative ecosystem in which tools, criteria, and human judgment interact. Students' concerns about fairness, misinterpretation, and accountability indicate an emerging awareness of assessment as a socially situated practice rather than a purely technical procedure (Henderson et al., 2019). This awareness resonates with critiques in the literature warning against the normalization of automated evaluation without critical engagement (Delikoura et al., 2025).

The post-assessment reflective task used in this study is inexpensive and easily replicated. Asking students to write briefly about their agreement with AI feedback and their views on AI as an evaluator requires no additional technology and can be embedded in any course that uses AI feedback tools. The reflective format appeared to surface reasoning that would not otherwise be visible to instructors, suggesting it merits wider adoption in AI-assisted assessment contexts

The findings tentatively suggest that post-assessment reflection could potentially mitigate some risks associated with GenAI over-reliance, though this hypothesis requires empirical testing. Rather than positioning GenAI feedback as an unquestioned authority, reflection appeared to encourage students to contextualize AI outputs, recognize their limitations, and reaffirm the role of instructor judgment. In doing so, reflection may help students understand not only how feedback operates but also how evaluative decisions are made and justified.

Figure 6 presents a conceptual model that synthesizes these findings by illustrating how transparent AI-based writing evaluation may initiate a reflective process among students. The model shows that AI-generated feedback can function as an initial instructional stimulus that, when disclosed to learners, may prompt critical reflection on both their written work and the evaluation mechanism itself. Through this reflection, students in this study moved beyond technical appraisal of the feedback to articulate awareness of AI's limitations and to distinguish feedback utility from evaluative authority. The model is offered as a theoretical synthesis of observed patterns rather than as a tested causal pathway.

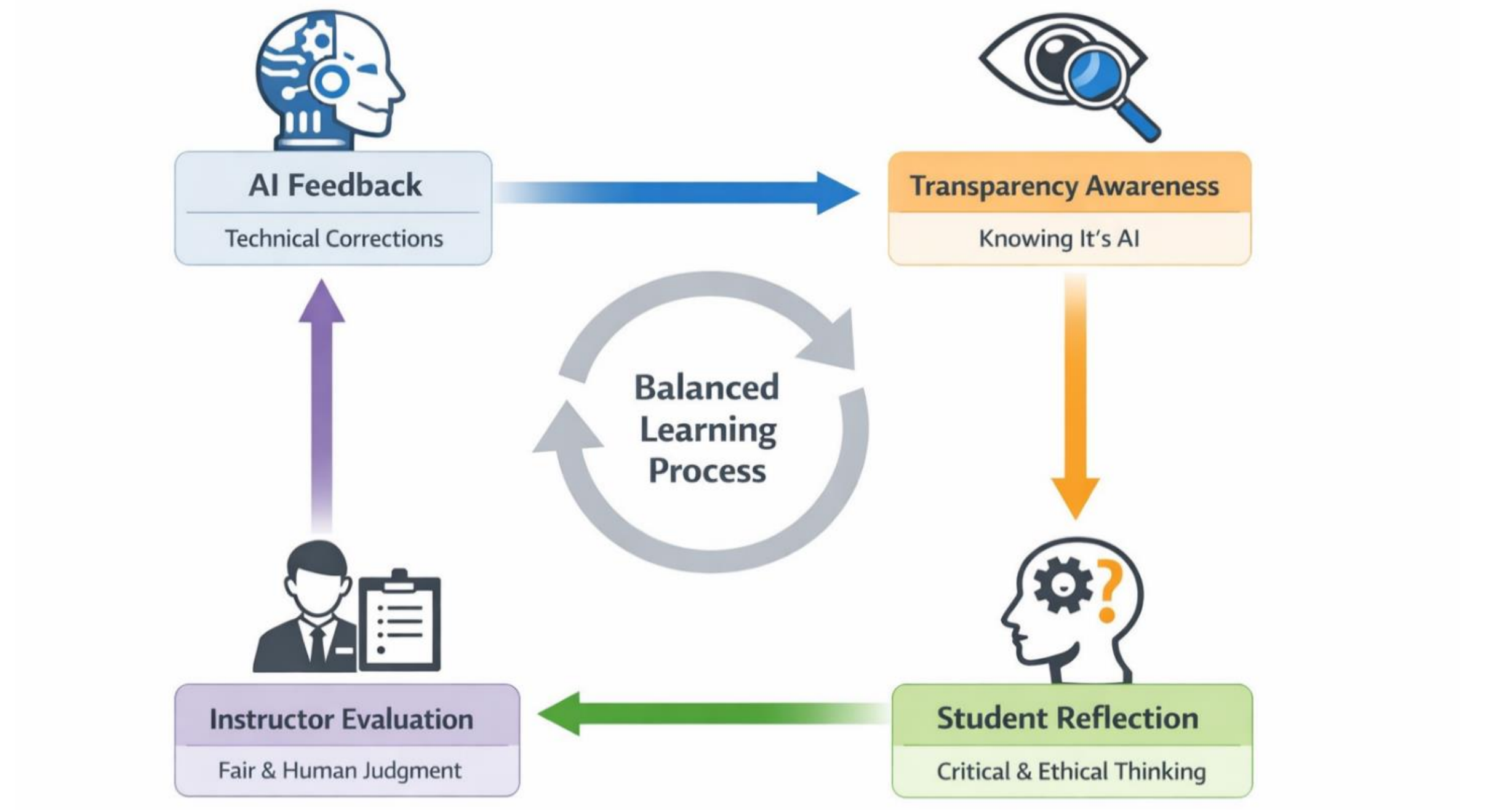

Figure 6. Conceptual model illustrating students' engagement with transparent GenAI-based writing evaluation.

## 6. Limitations and Future Work

This study has several important limitations that should be considered when interpreting the findings and that constrain the generalizability of the conclusions.

First, the analytic sample consisted of 13 students who completed reflective responses out of 19 enrolled, representing a small and potentially self-selected subset of the course population. This sample falls within the range commonly considered adequate for reflexive thematic analysis with a homogeneous, bounded sample (Guest et al., 2006; Braun & Clarke, 2021; Malterud et al., 2016), and thematic saturation appeared evident within the dataset, with all participants articulating Theme 1 and three of four themes recurring across more than half of the dataset. Nevertheless, saturation within a homogeneous sample does not establish that the same themes would emerge in samples with different characteristics.

The sample is also demographically homogeneous in important respects. All participants were male undergraduate computing students at a single Saudi public university, in a single course section, taught by a single instructor, during a single semester. This homogeneity restricts the transferability of findings to other populations, including women, students in non-computing disciplines, students at other institutions, and students from different cultural contexts.

Because the researcher served as both the course instructor and principal investigator, students may have provided responses they perceived as expected or favorable, despite assurances that reflections would not affect grades. The mitigation procedures adopted to address this risk are described in Section 3.5. Nevertheless, the possibility of response bias cannot be entirely ruled out.

The comparison between this transparent study and the earlier blinded study (AlGhamdi, 2024) is limited by several confounding factors. The two studies involved different student cohorts, were conducted in different academic years, and occurred against different levels of cultural and institutional familiarity with GenAI. In the earlier study, ChatGPT was relatively new, and student exposure was limited; by the time of the current study, GenAI tools had become normalized in higher education. These differences make it impossible to isolate the effect of transparency from cohort effects, temporal effects, or broader shifts in student attitudes toward AI.

The model presented in Figure 6 is offered as a theoretical synthesis of observed patterns rather than as a tested causal pathway. Empirical validation of the proposed stages, particularly the feedback loop from instructor authority back to AI integration, requires further research with longitudinal and comparative designs.

Future research should address these limitations by employing larger and more diverse samples, incorporating mixed-method or experimental designs with comparison groups, tracking perceptions longitudinally, and examining AI-mediated assessment across multiple disciplines, institutions, and cultural contexts. Studies conducted by researchers independent of the instructional context would also strengthen the credibility of findings. Of these directions, the most consequential for testing the conceptual model proposed in this study would be within-subject or experimental designs comparing student reflection under blinded and transparent evaluation conditions in the same cohort.

## 7. Conclusion

This study examined how students respond when they are explicitly informed that an AI system has evaluated their writing. Within the bounded context and sample of this study, participants' reflections suggest that disclosure of AI involvement prompted critical engagement with the evaluation rather than passive acceptance, with attention extending beyond the text itself to questions of authority, trust, and the role of human instructors. The present design cannot demonstrate that transparency caused this engagement; the pattern, however, was consistent across participant reflections. Across the dataset, ChatGPT-generated feedback was recognised for its clarity and usefulness, particularly in identifying surface-level writing issues. This perceived utility, however, did not translate into acceptance of GenAI as a legitimate or autonomous evaluator. Participants instead articulated a form of conditional trust in which AI was valued as a supportive feedback mechanism but not as a replacement for human oversight. Instructor judgement appeared to remain central to participants' perceptions of fairness, accountability, and trust in assessment. The central conceptual contribution of this study is the distinction between feedback utility and evaluative authority: two functions of AI in assessment that students appear to evaluate separately even within the same evaluation event.

Returning to the research questions, the study yields the following answers within the context of the sample and design. With respect to RQ1 (how students evaluate ChatGPT-generated feedback once they know ChatGPT produced it), participants regarded the feedback as clear, structured, and useful for surface-level revision, particularly for grammar, sentence structure, and organisation, while questioning the appropriateness of an AI system as the final evaluator of their work. With respect to RQ2 (how students reason about ChatGPT's role as an evaluator), participants drew a consistent distinction between feedback utility and evaluative authority: AI was accepted as a useful feedback mechanism but rejected as an autonomous grader, with participants emphasising the necessity of instructor review of AI outputs before any grading decision. With respect to RQ3 (the broader pedagogical and institutional concerns that emerge in students' reflections), participants raised concerns about fairness in light of potential AI errors, about the dialogical and individually attentive functions of human teaching that AI cannot replicate, and about the institutional purpose of higher education when assessment is delegated to AI.

By situating reflection after assessment, the study revealed a shift in students' focus from individual textual correction to broader systemic questions about evaluation practices. This pattern points to the potential role of reflective activities in helping students critically engage with automated assessment processes, a possibility that warrants further investigation. In this context, transparency served not only as an ethical consideration but also as a potential instructional strategy that supported critical engagement.

While this study is situated within a specific course context, its findings offer potentially transferable insights for educators navigating AI-mediated writing assessment. As GenAI becomes common in academic settings, productive responses are more likely to emerge from open and thoughtful integration than from prohibition. To support this goal, Figure 7 presents a proposed framework for integrating GenAI in writing assessment. The framework is organized around three guiding principles: transparency, human mediation, and reflective practice, with specific implications for instructors, assessment design, and students.

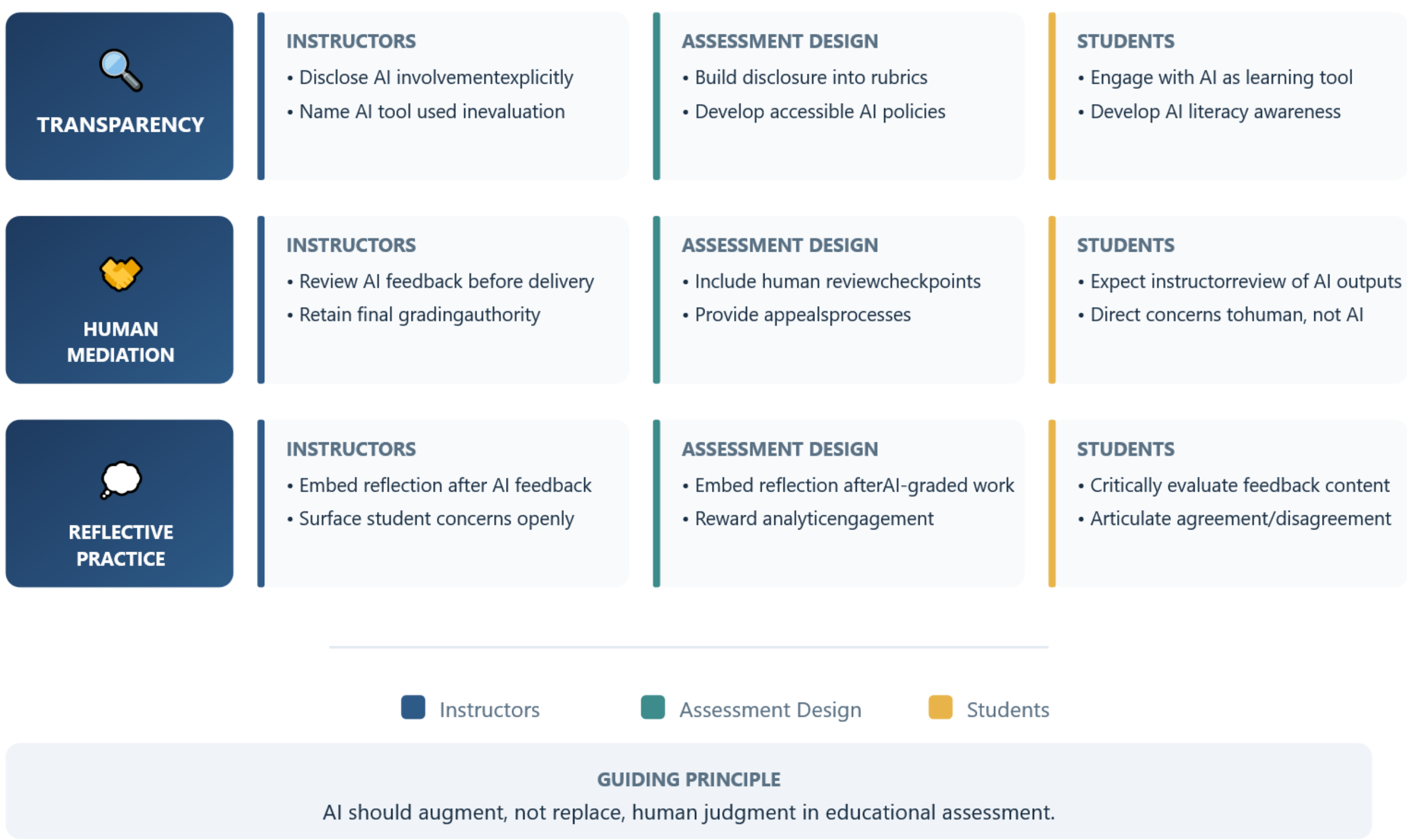

Figure 7. Proposed framework for ethical GenAI integration in writing assessment, derived from findings in this exploratory study.

As illustrated in Figure 7, transparency serves as the foundation, requiring clear communication about AI involvement. Human mediation ensures that instructors maintain evaluative authority, while reflective practice encourages critical engagement rather than passive acceptance. Together, these principles support an approach in which GenAI augments, rather than replaces, human judgment in educational assessment.

The implications for instructors using GenAI in assessment are tentative but suggestive. Disclosure about AI involvement, rather than undermining the assessment process, may create opportunities for productive reflection on authority, trust, and the conditions under which automated evaluation is appropriate. These observations, however, are drawn from a single course, a single semester, and thirteen participants. Studies across other disciplines, institutions, and cultural settings, and across different GenAI platforms, would be necessary before stronger conclusions could be advanced. As GenAI capabilities and student familiarity continue to evolve, the practical and ethical questions raised by AI-mediated assessment will require sustained empirical and pedagogical attention.

## Declaration Statements

### Data Availability Statement

The reflective-task prompts, coding schedule, theme definitions, and the Claude Code confirmatory-analysis session log supporting this study are available in an Open Science Framework repository at https://doi.org/10.17605/OSF.IO/S4HJ6. Anonymised transcribed reflections are included in the repository for research reuse. Original handwritten submissions are not shared because handwriting can be identifying even when names are removed.

### Use of AI:

*As a research instrument.* ChatGPT (OpenAI) was used as the AI evaluator in the pedagogical activity that constituted the study itself. Students' handwritten writing submissions were scanned and processed through ChatGPT, which generated rubric-based feedback and a numerical score as part of the course activity. This use of ChatGPT was the object of study, not an authorial aid; students' reflections on this ChatGPT-generated evaluation form the dataset analysed in this paper.

*As an analytic and editorial assistant.* Claude (Anthropic, model Opus 4.7) was used in two ways during the research process. First, Claude Code was used to read the folder of 13 scanned handwritten reflections and to conduct an independent confirmatory thematic analysis as a second analytic pass, following the inductive thematic analysis approach of Braun and Clarke (2006), as disclosed in Section 3.4; this use is documented in the methodology because it informed the refinement of the thematic structure. Second, Claude.ai was used to support language refinement and editorial work on the manuscript during writing. The author conceptualised the study, designed the methodology, conducted the primary NVivo-based analysis, made all interpretive and analytic decisions, and wrote and revised the manuscript.

Neither tool is listed as an author, and neither tool meets authorship criteria. The author retains full responsibility for the accuracy, originality, integrity, and conclusions of this work.